\documentclass[10pt,twocolumn]{article}

\newcommand{\Section}[1]{\vspace{-8pt}\section{\hskip -1em.~~#1}\vspace{-3pt}} 
\newcommand{\SubSection}[1]{\vspace{-3pt}\subsection{\hskip -1em.~~#1}
     	\vspace{-3pt}}

\newcommand{\ed}{\end{document}}

\newcommand{\hs}{\hspace}

\newcommand{\bit}{\begin{itemize}}
\newcommand{\eit}{\end{itemize}}
\newcommand{\ben}{\begin{enumerate}}
\newcommand{\een}{\end{enumerate}}

\newcommand{\beq}{\begin{equation}}
\newcommand{\eeq}{\end{equation}}
\newcommand{\ba}{\begin{array}}
\newcommand{\ea}{\end{array}}
\newcommand{\beqa}{\begin{eqnarray}}
\newcommand{\eeqa}{\end{eqnarray}}
\newcommand{\beqas}{\begin{eqnarray*}}
\newcommand{\eeqas}{\end{eqnarray*}}
\newcommand{\bfg}{\begin{figure}}
\newcommand{\efg}{\end{figure}}

\renewcommand{\a}{\alpha}

\begin{document}

\date{}

\title{
\Large\bf Fuzzy Rules and Evidence Theory for Satellite Image Analysis 
}

\author{ 
\begin{tabular}[t]{cc}
Arijit Laha & J. Das \\
National Institute of Management, Calcutta & Indian Statistical Institute, Calcutta \\
{\tt arijitl@yahoo.com} & {\tt jdas@isical.ac.in}
\end{tabular}
}

\pagenumbering{arabic}
\pagestyle{plain}
\setcounter{page}{1}
\maketitle
\pagestyle{empty}
\thispagestyle{empty}

\section*{\centering Abstract}

{\em
Design of a fuzzy rule based classifier is proposed. The performance of the classifier
for multispectral satellite image classification is improved using Dempster-Shafer theory
of evidence that exploits information of the neighboring pixels. The classifiers are tested
rigorously with two known images and their performance
 are found to be better than the results available in the literature. We also demonstrate
the improvement of performance while using D-S theory along with fuzzy rule based classifiers 
over the basic fuzzy rule based classifiers for all the test cases.
}

\Section{Introduction}
\label{sec:intro}

Analysis of satellite images has many
 important applications
such as prediction of storm and rainfall, estimation of natural
resources, estimation of crop yields, assessment of damage caused by natural disasters,
and land cover classification. In this paper we focus on land cover
classification from multi-spectral satellite images.

The most widely used techniques for this problem
employ
discriminant analysis, maximum likelihood classification, and neural networks 
 \cite{paola},  \cite{pinz}.  Such classifiers
 cannot handle the fact that for land cover a pixel may
correspond to more than one   types of objects. For example, the area covered
by a pixel may correspond to 30\% land and 70\% water. Note that, the uncertainty
involved in classifying such a pixel is not probabilistic, but fuzzy in nature
and  thereby it
demands  ``soft" classifiers.
In developing soft classifiers for land cover analysis two approaches have gained
popularity. These are  based on (1)fuzzy set theory   and (2) Dempster
and Shafer's (DS) evidence theory \cite{shafer}.

Numerous fuzzy classification techniques have been developed by many researchers to
solve problems in diverse fields. A comprehensive account of such works can be found in
 \cite{bez2}.    Fuzzy rules are attractive
because they are interpretable and
provides an analyst a deeper insight into the problem. Use of fuzzy rule based systems for
land cover analysis is   relatively new. In a recent paper B\'{a}rdossy
and Samaniego \cite{bard} have proposed a scheme for developing a fuzzy rulebased
classifier for analysis of multispectral images.

The other approach for designing soft classifiers is to use the evidence theory developed
by Dempster and Shafer \cite{shafer}.
 Since
the theory of evidence allows one to combine evidences obtained from diverse sources
of information in support a hypothesis, it seems a natural candidate for analyzing
multispectral images for land cover classification.

Here we propose a scheme for designing fuzzy rulebased classifiers
for land cover types that uses evidence
theory   for decision making. This is a two stage process.
First we find a good set of fuzzy rules using information from all channels.
In the next stage, the responses of the fuzzy rules over a $3\times 3$ neighborhood
are used to define 8 Basic Probability Assignment which are then combined by DS rule 
to exploit contextual
information to make a better decision. The problem
of high variation in the variances of different features, which often
degrades the performance of a distance based classifier substantially, is
  handled in a natural manner by fuzzy rules due to the atomic nature of the antecedent
clauses.

\Section{Designing the Fuzzy Rule base}

The proposed scheme has several stages. First a set of labeled prototypes is generated. Then
the prototypes are converted into fuzzy rules. The fuzzy rules are further tuned for
improving their performance. Labeled prototypes can be generated using any clustering
algorithm followed by labeling the cluster centers. However, for most of such algorithms
the number of clusters is a predefined parameter. Here we use the prototype generation scheme 
described in \cite{laha2}. It is a two stage algorithm involving unsupervised and supervised 
learning that dynamically decides the number of prototypes and extract them using the training
data. For details the
readers are referred to \cite{laha2}.

\SubSection{Designing the fuzzy rulebase}

A prototype (representing a cluster of points) ${\bf v}_i$ for   class $k$
can be translated into a fuzzy rule of the form :

\begin{center}
$R_{i}$ : $x_1$ is CLOSE TO $v_{i1}$ AND $\cdots$ AND $x_p$ is
CLOSE TO $v_{ip}$ then class is $k$.
\end{center}
The fuzzy set   CLOSE TO $v_{ij}$  is modeled by a
  Gaussian membership function :
$$\mu_{ij}  ( x_j; v_{ij},\sigma_{ij})=\exp{-{(x_j - v_{ij})}^2/{\sigma_{ij}}^2}.$$
Given a   data point ${\bf x}$ with unknown class, we first find the firing strength
of each rule. Let $\alpha_i({\bf x})$ denote the firing strength of the $i^{th}$
rule on a data point ${\bf x}$. We assign the point ${\bf x}$ to class $k$, if
$\alpha_r = \max_i(\alpha_i({\bf x}))$ and the $r^{th}$ rule represents class $k$.

Each fuzzy set is characterized by two parameters $v_{j}$ and $\sigma_{ij}$.
The $v_{ij}$s of the rules can be initialized with the components of the
final set of prototypes,$V^{final}$, generated by our SOFM based algorithm, $V^0= V^{final}=\{{\bf v}_1^{final}, \cdots, {\bf v}_{\hat c}^{final} \}$
$=\{{\bf v}_1^{0}, \cdots, {\bf v}_{\hat c}^{0} \}$
 where $v_{ij}^0=v_{ij}^{final}$. The notation $V^0$ is used to indicate that it corresponds
 to the initial centers of the membership functions.
 The initial estimates of the
$\sigma_{ij}$s are computed as follows.

For each prototype ${\bf v}_{i}^{0}$ in the set $V^{0}=\{{\bf v}_{i}^{0} \mid i=1,...,
\hat{c}, {\bf v}_{i}^{0} \in \Re^{p} \}$
let $X_{i}$ be the set of training data closest to ${\bf v}_{i}^{0}$. For each
${\bf v}_{i}^{0}$ the set
$$S_{i} = k_{w}\{ \sigma_{ij} \mid \sigma_{ij}
=(\sqrt(\sum_{{\bf x}_{k} \in
X_{i}}(x_{kj}-v_{ij}^0)^2))/|X_{i}|\}$$
is computed and is associated with the prototype. We use the $k_w \sigma_{ij}$ as the spread
of the membership function whose center is at $v_{ij}$;  $k_{w} > 0$ is a constant parameter
and its value can
have a significant
impact on the classification performance for complex data sets.

\SubSection{Tuning the rulebase}

The initial rulebase $R^0$ thus obtained is further refined to achieve better
performance. The exact tuning algorithm depends on the conjunction
operator  used for computation
of the firing strengths. The firing strength can be calculated using any
T-norm \cite{bez2}. Use of different T-norms
results in different classifiers. The {\bf minimum} and the
{\bf product} are among the most popular T-norms used as conjunction operators.
It is much easier to formulate a calculus based tuning algorithm if
product is used. However, if there are many clauses in the antecedent, the firing strength
of a rule tends to have low numerical values even when the membership value
of each individual clause is quite high. Though computationally this does not
pose any problem (we are interested in relative firing strengths of the rules),
it is conceptually somewhat unattractive - especially from the interpretability viewpoint.

Thus to avoid the use of the product and at the same time to be
able to   derive update rules easily we use a soft-min operator.

The {\bf soft-match} of $n$ positive number
$x_1,x_2,...,x_n$ is defined by
$$SM(x_1,x_2,...,x_n,q) = \left\{\frac{(x_1^q+x_2^q+...+x_n^q)}{n}\right\}^{1/q},$$
where $q$ is any real number.
$SM$ is known as an aggregation operator with upper bound of value 1 when
$x_i \in [0,1] \forall i$.
 It is
easy to see that
$\lim_{q\to \infty}SM(x_1,x_2,...,x_n,q) = \max(x_1,x_2,...,x_n)$
and
$\lim_{q\to -\infty}SM(x_1,x_2,...,x_n,q) = \min(x_1,x_2,...,x_n).$
Thus we define the softmin operator as the soft match operator with a sufficiently
negative value of the parameter $q$. The firing strength of the r-th rule
computed using softmin is

$$\alpha_r({\bf x}) = \left\{\frac{\sum_{j=1}^{j=p}(\mu_{rj}(x_j; v_{rj},
\sigma_{rj}))^q}{p}\right\}^{1/q}.$$
In the present study we use $q = -10.0$.

Let ${\bf x} \in X$
be from class $c$ and ${ R}_{c}$ be the rule from class $c$
giving the maximum firing strength
$\a_{c}$ for  ${\bf x}$. Also let ${R}_{\neg c}$ be the rule
 from
the incorrect classes having the highest firing strength
$ \a_{\neg c}$ for  ${\bf x}$.

We use the error function 
$ E = \sum_{{\bf x} \in X}(1-\a_{c}+\a_{\neg c})^{2}. $

We minimize $E$ with respect to $v_{cj}$, $v_{\neg cj}$ and $\sigma_{cj}$,
$\sigma_{\neg cj}$ of the two rules $R_{c}$ and $R_{\neg c}$ using gradient decent.
Here the index $j$ corresponds to clause number in the corresponding rule.
Minimizing $E$  will refine the rules with respect to their contexts in the {\it feature space}.
Note that, the context referred here is different from the context of
a pixel defined in terms of its spatial neighborhood.
 The tuning process is repeated until the rate of
decrement in E becomes negligible resulting in final rule base $R_{final}$.

\Section{Using the theory of evidence for Rule aggregation}

For the sake of completeness, we briefly introduce the
Dempster-Shafer theory of evidence.
Let $\Theta$ be the universal set and $P(\Theta)$ be its power set. A Belief 
measure is a function $Bel : P(\Theta) \rightarrow [0,1]$ that satisfies the axioms 
\cite{shafer}.\hfill\linebreak
$b1 :Bel(\emptyset) = 0$ and $Bel(\Theta) = 1$.\\
$b2 :$ For every $A, B\in P(\Theta)$, if $A \subset B$ then $Bel(A) \leq Bel(B)$.\\
$b3 :Bel(A_1\cup A_2\cup \cdots \cup A_n) \geq \sum_i Bel(A_i) - \sum_{i<j} Bel(A_i\cap A_j)
+\cdots +(-1)^nBel(A_1\cap \cdots \cap A_n)$, for every $n$ and for every collection
of subsets of $\Theta$.

There is a plausibility measure with each belief measure defined by
$Pl(A) = 1-Bel(A^c) \forall A \in P(\Theta)$.

Every belief measure and its dual plausibility measure can be expressed in terms of a
Basic Probability Assignment (BPA) function $m$. $m : P(\Theta) \rightarrow [0,1]$ is called
a BPA iff $m(\emptyset) = 0$ and $\sum_{A\subseteq \Theta} m(A) =1$. A belief
measure and a plausibility measure are uniquely determined by $m$ through the formulas:
\begin{equation} Bel(A) = \sum_{B\subseteq A}m(B).\end{equation}
\begin{equation} Pl(A) = \sum_{B\cap A\neq \emptyset}m(B) \hs*{2 mm}\forall A\subset \Theta.\end{equation}

Every set $A\in P(\Theta)$ for which $m(A) > 0$ is called a focal element of $m$.
Evidence obtained in the same context from two distinct sources and expressed by two
BPAs $m^1$ and $m^2$ on some power set $P(\Theta)$ can be combined by Dempster's rule of
combination to obtain a joint BPA $m^{1,2}$ as:
\begin{equation} m^{1,2}(A) = \left\{\begin{array}{ll}
            \frac{\sum_{B\cap C=A}m^1(B)m^2(C)}{1-K} &\mbox{if $A\neq \emptyset$}\\
            0 &\mbox{if $A=\emptyset$}
            \end{array}
        \right.\end{equation}
Here \[K = \sum_{B\cap C = \emptyset}m^1(B)m^2(C).\]

Eq. (3) is often expressed with the notation $m^{1,2}=m^1\oplus m^2$. The rule is
commutative and associative. Evidence from any number (say $k$) of distinct sources can be
combined by repetitive application of the rule as $m = m^1\oplus m^2\oplus\cdots\oplus m^k
=\oplus_{i=1}^km^i.$

\SubSection{Pignistic probability}

Given a belief measure we are often required to make decisions based on the available
evidence. In such case $\Theta$ becomes the set of decision alternatives and the function
$Bel$ denote our belief about the choice of the optimal decision $\theta_0 \in \Theta$.
However, in general it is not possible to select the optimal decision directly from the
evidence embodied in the function $Bel$. In such cases, we use the {\em pignistic transformation},  $\Gamma_\Theta$,
to construct a probability function for selecting the optimal decision  \cite{smets}.
Thus
$$ P^\Theta = \Gamma_\Theta (Bel).$$
$P^\Theta$ is called a {\em pignistic probability}, which can be used for making decision .
The pignistic probability for $\theta \in \Theta$ can be
expressed in terms of BPAs as follows:
\begin{equation} P^\Theta(\theta) = \sum_{A\subseteq \Theta, \theta\in A}\frac{m(A)}{\mid A\mid
}\end{equation}

Optimal decision can now be chosen in favor of $\theta_0$, if $\theta_0$ has the highest pignistic probability.

\SubSection{Scheme for decision making}

In our   problem the frame of discernment is the set of classes,
$\cal{C}$=$\{C_1,C_2,\cdots C_c\}$, where $c$ is the number of classes.
The propositions   take the form {\em
the true class label of the pixel of interest is in $A \subset \cal{C}$}.

Let us denote the pixel of interest as $p^0$ and its
eight spatial neighbors as $p^1, p^2,\cdots p^8$. We use the firing strengths produced
by the rulebase in support of different classes for $p^0$ and one of its neighbors, say
$p^i$ as the $i$-th source of evidence.
Let $r$ be the number of rules in the fuzzy rulebase. Since $c\leq r$, there could be
multiple rules corresponding to a class. Let $\a_k^0$ be the highest firing strength
produced by the rules corresponding to the class $C_k$ for $p^0$. We treat this value
as the confidence measure of the rulebase pertaining to the membership of $p^0$ to the
class $C_k$. Thus, the set of values $CM^0 = \{\a_k^0 : k=1,2,\cdots c\}$ contain the
confidence measures for all the classes for $p^0$ (if a confidence measure is less
than a threshold, say 0.01, it is set to 0). A similar set of confidence measures $CM^i$
can be constructed for every $p^i; i=1, \cdots , 8$.

Now we use $CM^0$ and $CM^i$ to define the $i$-th BPA $m^i$ to the subsets
of $\cal{C}$. There are $2^c$ possible subsets of $\cal{C}$, i.e.,  members of the
power set of $\cal{C}$. Each subset corresponds to the proposition that the ``true"
class of $p^0$ is contained in that subset. We shall consider the subsets
containing one and two elements only. The subsets containing one element correspond
to propositions of the form ``the class contained in the subset is the true class for
$p^0$" and the subsets containing two elements corresponds to propositions of the form
``the true class label of $p^0$ is any one of the two classes contained in the subset".
Assigning BPA to a subset essentially involves committing some portion of belief
in favor of the proposition represented by the subset. So the scheme followed for
assigning BPAs must reflect some realistic assessment of the information available
in favor of the proposition. We define $m^i$ as follows:

\begin{equation} m^i(\{C_k\}) = \frac{\frac{(\a_k^i + \a_k^0)}{2}\exp^{- (\a_k^i - \a_k^0)^2}}{S}, k=1,2,...,c\end{equation}

For $l,m=1,2,...,c,$ $m^i(\{C_l,C_m : l<m\}) = $
\begin{equation}\frac{\frac{(\a_l^i + \a_m^0)}{2}\exp^{- (\a_l^i - \a_m^0)^2}
            +\frac{(\a_m^i + \a_l^0)}{2}\exp^{- (\a_m^i - \a_l^0)^2}}{2S}\end{equation}
where $S = \sum_{k=1}^{k=c}\frac{(\a_k^i + \a_k^0)}{2}\exp^{- (\a_k^i - \a_k^0)^2} +$\\
$    \sum_{l=1}^{l=c-1}\sum_{m=l+1}^{m=c}\frac{(\a_l^i + \a_m^0)}{2}\exp^{- (\a_l^i - \a_m^0)^2}+$\\
$             \frac{(\a_m^i + \a_l^0)}{2}\exp^{- (\a_m^i - \a_l^0)^2}.$

The numerators in the right hand side of
the above formulae are measures of
confidence in favor of the respective propositions. A closer look on (5) shows that
the numerator is a product of two terms. The first term is the average of the confidence
measures of $p^0$ and $p^i$ for the class $C_k$, while the second term is an exponential
one that reflects the degree of closeness of the confidence measures. Thus as a whole
a high value of the numerator reflects two facts: (1) both $p^0$ and $p^i$ has high
confidence value for class $C_k$ and (2) the confidence values are close to each other.
Eq. (6) is a straightforward extension of the same concept when we define the
confidence in favor of a pair of classes.

Thus for the eight neighboring pixels we obtain eight combinable   sources of evidence.
The global BPA  can be computed by applying the Dempster's rule
repeatedly.  The combined
global BPA $m^G$ is computed as follows:
\begin{equation} m^G = \oplus_{i=1}^8m^i = (\cdots((m^1\oplus m^2)\oplus m^3)\oplus\cdots m^8). \end{equation}

It is easily seen that:

$m^{(i,j)}(\{C_k\})   =   m^i(\{C_k\})\oplus m^j(\{C_k\}) $

    $ =   \frac{\left\{\begin{array}{l}
            m^i(\{C_k\})m^j(\{C_k\})\\
            +m^i(\{C_k\})\sum_{l\neq k}m^j(\{C_k,C_l\})\\
            +m^j(\{C_k\})\sum_{l\neq k}m^i(\{C_k,C_l\})\\
            +\sum_{l\neq k}m^i(\{C_k,C_l\})\sum_{m\neq k,l}m^j(\{C_k,C_m\})
            \end{array}\right\}}{1-K}, $\linebreak
$k=1,2,...,c$

and
\begin{eqnarray}
m^{(i,j)}(\{C_l,C_m\}) & = & m^i(\{C_l,C_m\})\oplus m^j(\{C_l,C_m\}) \nonumber\\
    & = & \frac{m^i(\{C_l,C_m\})m^j(\{C_l,C_m\})}{1-K},
\end{eqnarray}
$l,m = 1,2,...,c,l\neq m$;
where $K$ is given by
\[ \begin{array}{lll}
    K& = &\sum_{k=1}^{c-1}m^i(\{C_k\})\sum_{l=k+1}^cm^i(\{C_l\})\\
    & &+\sum_{k=1}^cm^i(\{C_k\})\sum_{l,m\neq k}^cm^j(\{C_l,C_m\})\\
    & &+\sum_{k=1}^cm^j(\{C_k\})\sum_{l,m\neq k}^cm^i(\{C_l,C_m\})\\
    & &+\sum_{l\neq r,s, \:\hbox{and}\: m\neq r,s}^cm^i(\{C_l,C_m\})m^j(\{C_r,C_s\}).
    \end{array} \]

Once $m^G$ is obtained the pignistic probability
for each class is computed. The following formula is used for computing the pignistic
probability of class $C_k$:
\begin{equation} P^{\cal{C}}(C_k) = m^G(\{C_k\}) + \frac{\sum_{l=1,\:l\neq k}^c m^G(\{C_k,C_l\})}{2}\end{equation}

The pixel $p^0$ is assigned to the class $C_k$ such that
$$P^{\cal{C}}(C_k)\geq P^{\cal{C}}(C_l)\:\forall C_l \in \cal{C}.$$

\Section{Experimental results and discussions}

We report the performances of the proposed classifiers for two
multispectral satellite images. We call them {\bf Satimage1} and {\bf Satimage2}.

The  Satimage1 is a 256-level  Landsat-TM image of size $512 \times 512$ pixels captured by
seven sensors operating in different spectral bands. Each sensor generates
an  image with pixel values varying from 0 to 255. The $512 \times 512$
 ground truth data
provide the actual distribution of classes of objects captured in the image. From this
data we produce the labeled data set with each pixel represented by a 7-dimensional
feature vector and a class label.  
Satimage2 also is a seven channel 256-level  Landsat-TM image of size $512\times 512$. However
due to some characteristic of the hardware used in capturing the images the first
row and the last column of the images contain gray value 0. So we did not include
those pixels in our study and effectively worked with $511\times 511$ images. The
ground truth containing four classes is used for labeling the data. 

In our study we generated 4 training sets of samples for each of the images. For
Satimage1, each training set contains 200 data points randomly chosen from each of
eight classes. This choice is made to conform to the protocol followed in
\cite{kumar2}. For Satimage2 we include in each training set 800 randomly chosen
data points from each of four classes. Bischof et al. \cite{pinz}
used more training points / class than that of ours.

First we report the performances of the fuzzy rulebased classifiers using firing
strengths directly for decision making and compare the results with the published results.
Then we report the performances of the fuzzy  classifiers using evidence
theoretic approach for decision making. The performances of fuzzy rulebased classifiers
using firing strengths directly for decision making is summarized in the Table 1.

\begin{table}
{\scriptsize
\begin{center}
\begin{tabular}{|l|c|c|c|c|} \hline
Trng    &No. of     &$k_{w}$    &Error Rate in    &Error Rate in   \\
Set     &rules      &       	&Training Data    &Whole Image\\ \hline
\multicolumn{5}{|c|}{Satimage1}\\ \hline
1.      &30     &5.0        &12.0\%     &13.6\% \\ \hline
2.      &25     &6.0        &14.3\%     &14.47\% \\ \hline
3.      &25     &5.0        &12.0\%     &13.03\% \\ \hline
4.      &27     &4.0        &12.6\%     &12.5\% \\ \hline
\multicolumn{5}{|c|}{Satimage2}\\ \hline
1.      &14     &2.0        &16.3\%     &14.14\% \\ \hline
2.      &14     &2.0        &16.3\%     &14.04\% \\ \hline
3.      &12     &2.0        &17.09\%    &14.01\% \\ \hline
4.      &11     &2.0        &17.34\%    &14.23\% \\ \hline
\end{tabular}
 \end{center}
}
\vspace*{-0.5 cm}
\caption{Performances of fuzzy rulebased  classifiers using firing strength for
decision making for different training sets}
\vspace*{-0.4 cm}
\end{table}
For Satimage1 the best result reported in \cite{kumar2} uses a fuzzy integral based method
and gives the classification rate 78.15\%. In our case, {\em even the worst result is
about 5\% better than that}.

For Satimage2 the reported result in \cite{pinz} shows
84.7\% accuracy with the maximum likelihood classifier (MLC) and 85.9\% accuracy with
neural network based classifier.
In our case for all training-test partitions the fuzzy rulebased classifiers outperform the MLC
and at par with the results reported for neural networks.

Tables 2 summarizes the performances of the fuzzy rulebased classifiers using
evidence theoretic approach. We used the same
set of fuzzy rules as used previously, but the rule outputs are
aggregated using the evidence theory.

\begin{table}
{\scriptsize
\begin{center}
\begin{tabular}{|l|c|r|} \hline
Training    &No. of     &Error Rate in \\
Set     &rules      &Whole Image \\ \hline
\multicolumn{3}{|c|}{Satimage1}\\ \hline
1.      &30     &12.3\% \\ \hline
2.      &25     &13.37\% \\ \hline
3.      &25     &11.6\% \\ \hline
4.      &27     &11.03\% \\ \hline
\multicolumn{3}{|c|}{Satimage2}\\ \hline
1.      &14     &12.7\% \\ \hline
2.      &14     &12.65\% \\ \hline
3.      &12     &12.4\% \\ \hline
4.      &11     &12.51\% \\ \hline
\end{tabular}
 \end{center}
}
\vspace*{-0.5 cm}
\caption{Performances of the evidence theoretic fuzzy    classifiers    for
different training sets}
\vspace*{-0.5 cm}
\end{table}

Comparison of Table 2 with Table 1 clearly shows that in every case there
is a consistent improvement in the classification performance. In case of Satimage1 the
improvements varied between 1.1\% and 1.5\% and the best performing classifier (for training
set 4) achieves error rate as low as 11.03\%. For Satimage2 also the improvement varied
between 1.4\% and 1.7\%. So the overall improvement for Satimage1 over the existing methods
is more than 7\%. For Satimage2 also we achieved consistent improvements using training sets of smaller
size. For applications like crop yield
estimation even a small improvement will have a significant impact on the overall estimate.

\Section{Conclusion}
We proposed two classifiers: one is fuzzy rule based and the other integrates outputs of fuzzy rules
using theory of evidence.  Fuzzy rules are extracted with the help SOFM. The system automatically decides
on the number of rules.

The fuzzy rule based classifier is of general nature and can be applied in any classification
problem, while the evidence theoretic classifier exploits the spatial information
available for an   image to make the classification decision.

In the evidence theoretic framework we use the pixel under consideration and one of its neighbors
to provide a body of evidence in support of different propositions regarding the class membership
(to a particular class as well as a pair of classes) of the pixel. The BPAs for the propositions
are calculated from the mutual confidences of the pixels in support of respective propositions.
Eight bodies of evidence is obtained for eight neighbors of the pixel. Now the evidences are
combined  to obtain a global body of evidence. Then pignistic
probability for each class is computed and the pixel is assigned to the class with highest
pignistic probability. The proposed system demonstrates a consistent improvement
in performance.

{\bf Acknowledgement:} Authors thank Prof. N. R. Pal for his continuous support and advices.
They also thank Dr. A. S. Kumar and Dr. A. J. Pinz for allowing them
to use the satellite images Satimage1 and Satimage2 respectively to test the proposed methods.

\ed